\documentclass[fleqn,10pt]{wlscirep}
\usepackage[utf8]{inputenc}
\usepackage[T1]{fontenc}
\usepackage{algorithm}%
\usepackage{algorithmicx}%
\usepackage{algpseudocode}
\usepackage{array}
\usepackage[export]{adjustbox}
\usepackage{multirow}%
\usepackage{hyperref}

\algrenewcommand\algorithmicrequire{\textbf{Input:}}
\algrenewcommand\algorithmicensure{\textbf{Output:}}
\algnewcommand\algorithmicforeach{\textbf{for each}}
\algdef{S}[FOR]{ForEach}[1]{\algorithmicforeach\ #1\ \algorithmicdo}
\let\oldReturn\Return
\renewcommand{\Return}{\State\oldReturn}

\newcommand\MyBox[2]{
  \fbox{\lower0.75cm
    \vbox to 1.7cm{\vfil
      \hbox to 1.7cm{\hfil\parbox{1.4cm}{#1\\#2}\hfil}
      \vfil}%
  }%
}
\newcolumntype{P}[1]{>{\centering\arraybackslash}p{#1}}

\title{SAILOR: Perceptual Anchoring For Robotic Cognitive Architectures}

\author[1,*]{Miguel Á. González-Santamarta}
\author[1]{Francisco J. Rodrıguez-Lera}
\author[1]{Vicente Matellan-Olivera}
\author[1]{Virginia Riego Del Castillo}
\author[1]{Lidia Sánchez-González}
\affil[1]{Robotics Group, Department of Mechanic Engineering, Computer and Aerospacial Sciences, University of León, León, 24006, Spain}

\affil[*]{mgons@unileon.es}

\begin{abstract}
Symbolic anchoring is a crucial problem in the field of robotics, as it enables robots to obtain symbolic knowledge from the perceptual information acquired through their sensors. In cognitive-based robots, this process of processing sub-symbolic data from real-world sensors to obtain symbolic knowledge is still an open problem. To address this issue, this paper presents SAILOR, a framework for providing symbolic anchoring in the ROS 2 ecosystem. SAILOR aims to maintain the link between symbolic data and perceptual data in real robots over time,  increasing the intelligent behavior of robots. It provides a semantic world modeling approach using two deep learning-based sub-symbolic robotic skills: object recognition and matching function. The object recognition skill allows the robot to recognize and identify objects in its environment, while the matching function enables the robot to decide if new perceptual data corresponds to existing symbolic data. This paper provides a description of the proposed method and the development of the framework, as well as its integration in MERLIN2 (a hybrid cognitive architecture fully functional in robots running ROS 2).
\end{abstract}
\begin{document}

\flushbottom
\maketitle
%
%
\thispagestyle{empty}


\section*{Introduction}

The use of cognitive architectures~\cite{ye2018survey,kotseruba202040,kotseruba2016review} as a mechanism for generating robot behaviors is broadly accepted. There are three types of cognitive architectures: symbolic, based on the principles of symbolic AI that include knowledge representation, reasoning, and planning modules; emergent, based on the principles of emergence and complex systems where behavior emerges from interactions between components; and hybrid, a combination. Although there is no well-defined degree of hybridization between symbolic and emergent concepts, usually they work with symbolic information, using for instance the PDDL (Planning Domain Description Language)~\cite{fox2003pddl2}.

In the ROS community, PDDL is highly accepted. There are several PDDL-based tools used in practical cognitive architectures, however, two of them are well known: ROSPlan~\cite{cashmore2015rosplan} and PlanSys2~\cite{martin2021plansys2}. Both apply PDDL to represent the real world of the robot by creating the objects and the attributes of the world. Afterward, it is used by symbolic planners such as POPF (Partial Order Planning Forwards)~\cite{popf} to generate plans to achieve the goals of the robots. 

How the world model serves the architecture and how is updated over time is the problem here. As a result, the research question faced in this paper is how to perform the task of creating and maintaining the correspondences between symbolic data and sensor data in a cognitive architecture for robots.

Obtaining PDDL knowledge from raw perceptual data is not straightforward. Thus, perception is the process of converting raw sensory data into cognitive architectures' internal representation, particularly symbolic knowledge. Whereas symbolic anchoring~\cite{coradeschi2003introduction} is the process of creating and maintaining the correspondence between symbols and sensor data that refer to the same physical objects. Not only knowledge creation is needed, but also knowledge maintenance. This is an aspect of the Symbolic Knowledge Grounding~\cite{harnad1990symbol} that is the problem of how to ground the meanings of symbols used by the robot. The process of grounding symbols to real-world objects by a physical agent interacting in the real is known as Physical Symbol Grounding~\cite{vogt2002physical}.

This paper presents SAILOR (Symbolic AnchorIng from perceptual data for rOs2-based Robots) a component that creates and maintains real-time knowledge in a robot. SAILOR is integrated into a hybrid cognitive architecture that acquires data from the environment through sensors and automatically builds knowledge from it. To do that, it combines two sub-symbolic skills: one to obtain information by recognizing the objects that the robot sees and the other one to decide if there is new knowledge or if old knowledge has to be updated.

SAILOR can be integrated into a ROS 2-based hybrid cognitive architecture improving the learning capability of the robot because the robot can learn new symbolic knowledge. 
As a result, knowledge about the real world is obtained in real-time as the robot interacts with the world rather than only using innate knowledge manually created before running the cognitive architectures, providing the robot with the ability to increase its knowledge about the world autonomously. Besides, this knowledge can be obtained from the effects of well-known actions, such as navigation, whose final state can be inferred using a symbolic planner.

\subsection*{Contributions and article overview}

The main contribution of this paper consists of a framework to perform symbolic anchoring in ROS 2. In more detail, this research provides the following contributions:

\begin{enumerate}
    \item An updated symbolic anchoring pipeline based on state-of-the-art works. Our approach is based on object detection followed by physical feature extraction based on point cloud, which differs from the literature approaches based on point cloud segmentation followed by image classification. This improves object detection and reduces the computational complexity of processing the full point cloud.
   
    \item A matching function based on object tracking and deep learning for a symbolic anchoring system, which provides information on whether an object is the first time the robot has seen it.
    
    \item Evaluation of the existing solutions for symbolic anchoring with the state-of-the-art indoor and outdoor datasets. 
    
     \item Implementation and validation of SAILOR proposed solution inside a cognitive architecture for ROS 2.
\end{enumerate}


\section*{Related Works} 
\label{sec:related_works}

Symbolic anchoring is the process of creating and maintaining the link between symbolic data and sensor data. Symbolic anchoring systems are based on extracting features from physical objects. Then, they are used to check if new perceived objects correspond with known objects. This mechanism is commonly known as the matching function. As a result, the symbolic anchoring problem~\cite{coradeschi2003introduction} is related to how to perform symbolic anchoring in an artificial intelligence system. In fact, symbolic anchoring is a special case of Physical Symbol Grounding~\cite{vogt2002physical} where symbolic data is maintained and updated in time. The symbolic anchoring problem involves different areas, such as psychology, cognition, linguistics and computer science. This makes symbolic anchoring a complex process.

There are initial works in symbolic anchoring that use fuzzy logic to implement the matching function and the grounding, such as \cite{coradeschi1999anchoring}. In later works, we can find one of the basic symbolic anchoring pipelines presented in \cite{daoutis2009grounding}. It shows a basic system for symbolic anchoring using only visual features to implement the matching function used to check if new objects have to be stored in the knowledge base. This system was composed of a perceptual layer, where sensory data is generated and processed to extract features; an anchoring layer, where the grounding and symbolic anchoring processes take place; and a knowledge representation layer, which contains a knowledge base. This approach uses SIFT (Scale-invariant feature transform)~\cite{lowe2004distinctive} to produce visual features to check if new objects are obtained. The grounding is based on describing the physical objects using color, object class, semantic localization and spatial relations.

The work~\cite{elfring2013semantic} presents a method for modeling the semantic environment of a robot using probabilistic multiple hypothesis anchoring (PMHA), which includes the matching function. The method uses probabilistic reasoning to update the robot's understanding of the environment as it receives new sensory information, however, there is not a clear symbolic anchoring architecture of the complete system. The features used as input for its algorithm are color images and object shapes. It also presents a grounding skill that describes the objects using their size and their color. 

Another alternative is presented in \cite{persson2017learning}. This work proposed a method for improving the symbolic anchoring of objects in the real world through the use of learned actions. The feature used in the matching function is the Euclidean distance between two objects and the classification coefficient. Both features are used in the matching function that is based on a formula with a threshold for each input. Besides, the presented framework is based on integrating actions and perception to improve the accuracy of symbolic anchoring in situations where visual perception alone is unreliable. The framework is provided with machine learning models that learn actions that can be used to disambiguate objects thus improving the symbolic anchoring of objects in the real world. Another case of using object poses is presented in \cite{gunther2018context}. In this work, the features used are again the Euclidean distance and classification coefficient. These two features are fed into the matching function that is based on a Support Vector Machine (SVM) plus the use of the Hungarian Method~\cite{kuhn1955hungarian} to assign each perception to an existing anchor. The SVM is trained with a dataset created by the authors using a mobile robot.

More complex matching functions can be found in symbolic anchoring systems. In cite~\cite{persson2019semantic}, authors present a matching function based on machine learning techniques. The data used to train the model is composed of five similarities: object class, color histogram, distance, size and timestamp. The dataset was collected indoors using a fixed camera that was in front of the used robot manipulator. Then, several models are trained, such as SVM and KNN.

According to the literature, the use of symbolic anchoring in robots is still a problem to be faced. There are different approaches but all of them have in common the feature extraction from sensory data and the implementation of a matching function that allows knowing if new perceptions correspond with known objects. In this work, we present a new symbolic anchoring from perceptual data, based on ROS 2, that uses deep learning to carry out the matching function.

In computer vision, there is a wide variety of architectures available for object detection \cite{amjoud2023object}. It is approached from three points of view: the neural networks backbone such as AlexNet~\cite{krizhevsky2017imagenet} or ResNet~\cite{he2016deep}, detectors based on two-stage anchors such as Faster R-CNN~\cite{ren2015faster} or one-stage anchors such as YOLO~\cite{redmon2016you}, and detectors based on transforms such as DETR~\cite{carion2020end}. However, YOLOv8~\cite{Jocher_YOLO_by_Ultralytics_2023} has become a leading reference, since the software developed by Ultralytics allows fast, accurate and easy detection, classification, and segmentation.

Moreover, some tasks in the computer vision field have similarities with symbolic anchoring from the perceptual data. These tasks are object tracking~\cite{yilmaz2006object} and image retrieval~\cite{datta2008image}. Symbolic anchoring systems can be improved by using object tracking, as it locates and follows a particular object over a sequence of frames. Symbolic anchoring also requires extracting high-level symbolic representations from sensory data to decide whether something exists in current knowledge. In this sense, it is similar to image retrieval in that it searches for images in a database that are similar to a given image by extracting certain features and applying machine learning algorithms to find a match. However, symbolic anchoring from perception involves the extraction of high-level symbolic representations from sensory data.


One popular technique used for computing image similarity is the Siamese Convolutional Neural Networks~\cite{melekhov2016siamese}. These networks learn a similarity metric by training on pairs of images. This approach has shown promising results in various image retrieval applications \cite{qi2016sketch,rahman2022product,sharma2022siamese,zhang2022content}, allowing for effective comparison and matching of images based on their visual content.

The approach proposed in this paper creates a symbolic understanding of the observed environment by defining meaningful symbols of objects and their relationships. It goes beyond visual recognition as it offers a more abstract and interpretable representation for reasoning and decision-making in robotics. While image retrieval is focused on matching and categorization, symbolic anchoring goes further, extracting symbolic meaning and updating the existing knowledge, which enables robots to reason about their surroundings in a more semantically rich manner.

\section*{SAILOR Proposal}
\label{sec:materials_methods}

The Materials and Methods section of this paper aims to detail the components and processes used to develop a symbolic anchoring from perception system for robots. This system leverages the principles of cognitive architecture to allow robots to anchor their perceptions to symbolic representations, enabling them to process and understand the environment in a more human-like manner. The section is divided into two main subsections: Formalization of Symbolic Anchoring from Perceptual Data and SAILOR pipeline.
These subsections will provide a comprehensive understanding of the methodology and procedures used to build the symbolic anchoring system, as well as the datasets used.

\subsection*{Formalization of Symbolic Anchoring from Perceptual Data}

Symbolic anchoring is the task of creating and maintaining in time the correspondences between symbolic data and sensor data, also called percepts. Following \cite{coradeschi2003introduction,daoutis2009grounding,coradeschi2000anchoring,coradeschi2001fuzzy}, the symbolic anchoring is composed of three systems:

\begin{itemize}
    
    \item \textit{Perceptual System}: This system is in charge of generating percepts from the data obtained from the real world. It includes a set of percepts, $\Pi = \{\pi_1, \pi_2,...\}$. A percept is a data structure, $\pi_i$, that defines a physical object. Each percept has a set of measurable features, $\phi_i$ with values in the domain $D(\phi_i)$. As a result, the perceptual system includes a set of features, $\Phi = \{\phi_1, \phi_2,...\}$, used to describe each percept.

    \item \textit{Anchoring System}: This is the system in charge of updating the symbolic knowledge using percepts. This correspondence is represented by the data structure called anchor, $\alpha_i$. Thus, this system has a set of anchors, $A = \{\alpha_1, \alpha_2,...\}$.

    \item \textit{Symbolic System}: This system is in charge of maintaining symbolic knowledge and using it to reason about the actions needed to achieve certain goals. Symbolic knowledge is stored in a knowledge base composed of four sets:
    \begin{itemize}
        \item Set of classes $C = \{c_1, c_2,...\}$ that describes the classes of objects that can appear in the problem.
        \item Set of objects $O = \{o_1, o_2,...\}$ that contains the objects of the problem.
        \item Set of predicates $P = \{p_1, p_2,...\}$ that contains the attributes of the world.
        \item Set of facts $F = \{f_1, f_2,...\}$ that described the the world.
    
    \end{itemize}
    
\end{itemize}

Symbolic anchoring also has a \textit{predicate grounding relation} $G \subseteq P \times \phi \times D(\phi)$. This relation is in charge of encoding features $\phi_i$ from each $\pi_i$ using the properly predicates $p_i$ to create the facts $f_i$. On the one hand, percepts can be described using visual features, such as color histograms, descriptors and semantic object categories. On the other hand, \cite{gunther2018context} uses physical features, such as 3D position, 3D size and orientation. Finally, a combination of both, visual and physical features, can be used, as in the case of \cite{persson2019semantic}.

\subsubsection*{The functionalities of Symbolic Anchoring}

In the symbolic anchoring process, anchors $\alpha_i$ can be created both top-down and bottom-up. Bottom-up approaches are based on events from the perceptual system (e.g. new percepts obtained $\pi_i$ from object recognition) whose data can be linked to existing anchors. Nevertheless, top-down takes place when symbolic data needs to be related to a percept.

The maintenance of anchors takes place at each cycle of SAILOR when new percepts are created. The new percepts are compared with the existing anchors and two different situations can occur. Those percepts that match an anchor are used to update the anchor and the symbolic data with the new information. On the contrary, if there is no match, new anchors are created. To check if a percept matches an anchor, a matching function $M$ (Eq. ~\ref{matching_function}) is used. This function takes as input a percept $\pi_i$ and an anchor $\alpha_i$ and returns the degree of matching. 

\begin{equation}
    M : \pi \times \alpha \rightarrow [0, 1]
    \label{matching_function}
\end{equation}

There are four main processes associated with symbolic anchoring as presented in the literature (\cite{persson2019semantic}, \cite{loutfi2005maintaining}):

\begin{itemize}
    \item \textit{Acquire}: This process initiates new anchors whenever new percepts are received and do not match any existing anchor. It takes each new percept $\pi_i$ and each existing anchor $\alpha_i$ and computes the matching degree using the matching function $M$. For each percept that does not match an anchor, symbolic data (objects and facts) is created using the \textit{predicate grounding relation} $G$.

    \item \textit{Find}: This process takes an object $o_i$ and the facts $f_i$ that describe that object and returns an anchor $\alpha_i$. Then, that anchor is compared against the existing anchors and current percepts. If there is an anchor that matches, that anchor is selected. A new anchor is created if there is no match.

    \item \textit{Re-acquire}: This process is intended to extend the definition of a matching anchor $\alpha_i$ from the timestamp $\tau$ to timestamp $\tau + \kappa$. It is based on taking the matching percepts and updating the anchors and the symbolic data over time.

    \item \textit{Track}: This process is based on taking an anchor $\alpha_i$ defined in timestamp $\tau$ and extends its definition to timestamp $\tau + \kappa$. This can be performed by using the re-acquire functionality if a match takes place or by predicting the future state of the anchor after some elapsed time from the last observation. 
\end{itemize}

\subsection*{Pipeline}

\begin{figure}[!ht]
\centering
\includegraphics[width=0.65\textwidth]{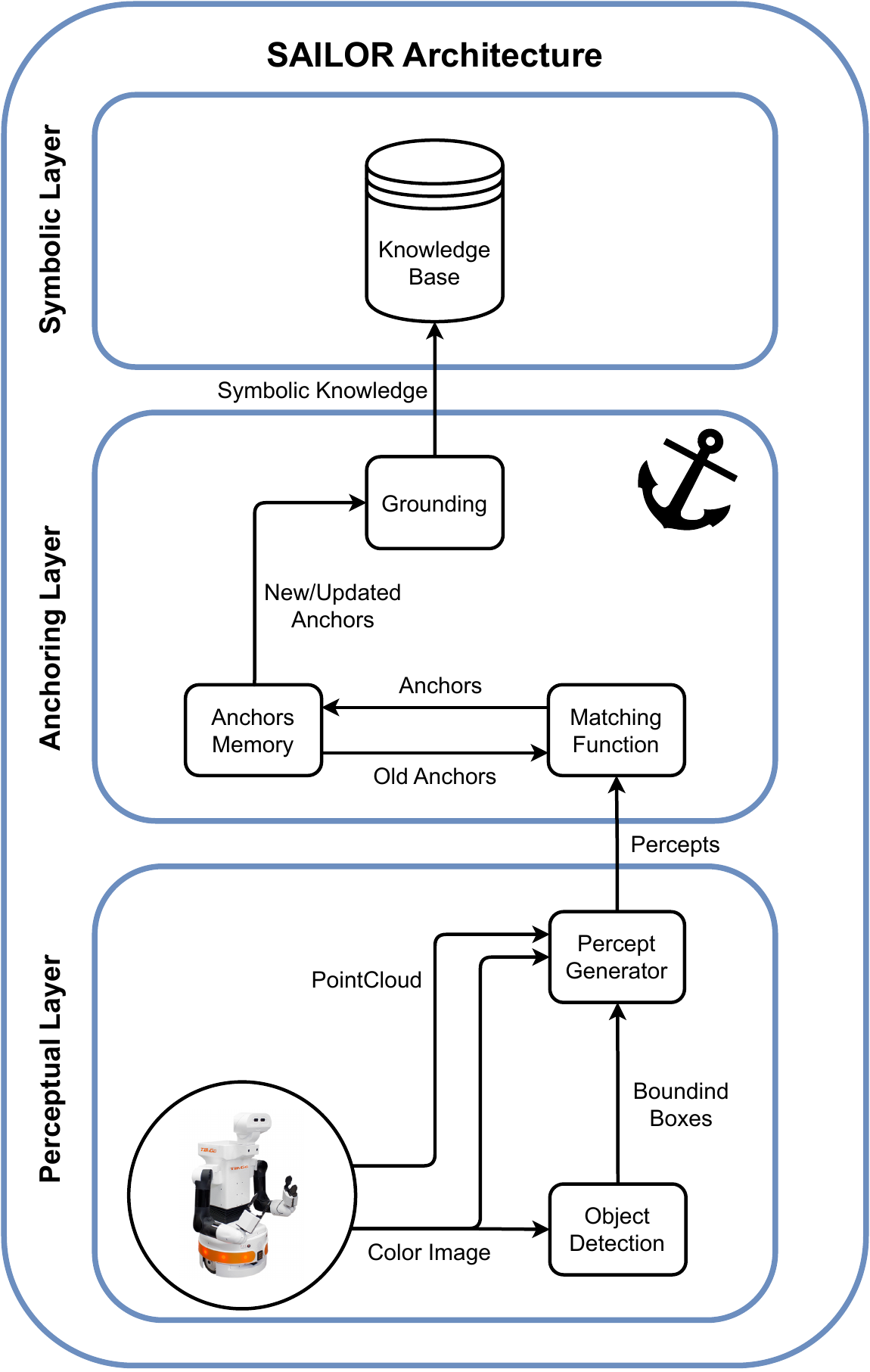}
\caption{\label{fig:sailor_architecture} SAILOR Architecture. It is composed of Symbolic Layer, Anchoring Layer and Perceptual Layers.}
\end{figure}

SAILOR (Symbolic AnchorIng from perceptuaL data for rOs2-based Robots) is the symbolic anchoring system integrated into the cognitive architecture. SAILOR's architecture, which is a bottom-up symbolic anchoring architecture, is presented in Figure~\ref{fig:sailor_architecture}. It is divided into three layers: perceptual, anchoring and symbolic layers as described in \cite{daoutis2009grounding}.

\subsubsection*{Perceptual Layer}
This is the layer that generates the percepts,  the correspondences between symbolic data and sensor data. To do this, an RGB-D camera is used. The color images captured from the camera are processed in an object detection system, which is a sub-symbolic skill of the robot. Then, a set of features of each percept is extracted from the color images, the obtained bounding boxes of the objects and the point clouds captured from the camera. Considering the features employed by the existing solutions analyzed in the related works section, we have selected the following five visual and physical features to describe each percept:
    \begin{itemize}
        \item Class of the object: this feature is obtained from the object detection skill. In this work, YOLOv8 (YOLOv8m)~\cite{Jocher_YOLO_by_Ultralytics_2023} was used to process color images obtained from the robot to detect objects in the environment since it was the best object detection system when this work took place.
        
        \item Tracking ID: this feature is obtained by applying the tracking algorithm ByteTrack~\cite{zhang2022bytetrack} to the output of YOLOv8. The default configuration of ByteTrack is employed.
        
        \item Cropped image: this feature is the image obtained after cropping the original frame from the camera using the bounding box detected by YOLOv8. 
        
        \item Position of the object: this feature is the object centroid in a 3D space of the detected object in meters. 

        From the obtained bounding box, the pixel location of its centroid (cx,cy) is determined. Next, these pixels are converted to the 3D camera coordinate system by using the obtained point cloud. Then, this position is transformed from the local coordinate system of the camera to the global coordinate system of the robot.
        
        \item Size of the object: this feature is the size of the object, in meters, represented as a box. The measurements are obtained by calculating the maximum and minimum, for each 3D axis, in the point cloud data within the limits of the bounding box. The data of the point cloud, the points, has to be aligned with the data of the image, the pixels.
        
        \item Timestamp: this feature is the timestamp of the obtained image.
    \end{itemize}

\subsubsection*{Anchoring Layer}
This is the layer in charge of managing the anchors, this means anchor robot perceptions to symbolic representations. Thus, it applies the matching function to each percept received from the perceptual layer. If these percepts, which belong to a certain frame, are new (they do not match an existing anchor), new anchors are created. Those percepts that match an anchor provide updated information to that anchor and its corresponding symbolic knowledge. This implies applying the acquire and reacquire functionalities respectively. 

The symbolic anchoring procedure is presented in Algorithm~\ref{alg:anchoring}. In the initial case, when there are no anchors, all received percepts are acquired, which involves anchor creations. In the following cases, the new percepts are used to create a matching table which is a matrix that is composed of the probabilities of each percept (rows) to match each anchor (columns).


\begin{algorithm}[ht!]
\caption{SAILOR's symbolic anchoring algorithm.}\label{alg:anchoring}
\begin{algorithmic}[1]
\Require $new\_percepts$
\If{$anchors == \emptyset$}
    \ForEach {$p \in new\_percepts$}
        \State $acquire(p)$
    \EndFor
\Else
    \State $M \gets create\_matching\_table(new\_percepts)$
    \State $H \gets hungarian\_method(M)$

    \ForEach {$i \in M$}
        \If{$i \in H$}
            \If{$H(i) > threshold$}
                \State $reacquire(p)$
            \Else
                \State $acquire(p)$
            \EndIf
        \Else
            \State $acquire(p)$ 
        \EndIf
    \EndFor
\EndIf
\end{algorithmic}
\end{algorithm}

The procedure to create the matching table is shown in Algorithm~\ref{alg:table}. It is based on applying the matching function to each candidate, that is each new percept and to the existing anchors. Each pair of percept-anchor is compared using their features. As a result, the table obtained is a matrix $N\times M$, where $N$ is the number of new percepts and $M$ is the number of existing anchors. Each cell contains the matching value for each pair of percept-anchor which is a value in the range [0, 1].

Then, the Hungarian Method~\cite{kuhn1955hungarian} is applied to the matching table. This algorithm is used to solve the assignment problem, which involves finding the optimal assignment of agents to tasks in a cost-minimization or profit-maximization scenario. The output is the rows and columns that correspond to the associations. As a result, each percept is associated with its corresponding anchor. 

Finally, the pairs of percept-anchor of the associations with a matching value greater than a threshold, that is 0.5, are the reacquire cases. However, the pairs with a value below the threshold are acquiring cases. There are also acquiring cases when new percepts are not part of the result of the Hungarian Method. This can happen if the number of new percepts is greater than the number of existing anchors.

\begin{algorithm}[ht!]
\caption{Create matching table algorithm.}\label{alg:table}
\begin{algorithmic}[1]
\Require $new\_percepts$
\Ensure $m\_table$
\State $m\_table \gets \emptyset$
\ForEach {$p \in new\_percepts$}
    \State $m\_row \gets \emptyset$
    \ForEach {$a \in anchors$}
        \State $m\_value \gets matching\_function(p, a)$
        \State $m\_row \gets m\_row \cup m\_value$
    \EndFor
    \State  $m\_table \gets m\_table \cup m\_row$
\EndFor
\Return $m\_table$
\end{algorithmic}
\end{algorithm}

\subsubsection*{Matching Function}
 
The matching function implemented in the presented solution checks
if new perceived objects correspond with known objects (stored percepts).  
Thus, a match is obtained if each pair of percept-anchor tracking IDs is the same as the new one. 

The comparison of each pair of percept-anchor gives the input features of the neural network which are the following:
\begin{itemize}
    \item Same object class: this feature is a boolean that indicates if the classes of the percept and the anchor are the same.
    \item Cropped images: these RGB images correspond to the cropped images of the percept and the anchor. They are used to measure visual similarity. The images are preprocessed following the next steps:
    \begin{itemize}
        \item The images are cropped to a size of 224x224 pixels. This means that a square region of the image is selected, discarding the remaining parts.

        \item Then, they are resized to a size of 256x256 applying bilinear interpolation, which is a technique to estimate pixel values in the resized image based on the surrounding pixels. Bilinear interpolation helps in preserving the visual quality of the images during the resizing process. This step ensures that all the input images have the same size for consistency.

        \item The pixel values of the images are normalized using mean and standard deviation values of ImageNet, which are [0.485, 0.456, 0.406] and [0.229, 0.224, 0.225] respectively. This normalization centers the pixel values around zero, which helps in reducing the effect of lighting variations in the images; and scales the pixel values to have a standard deviation of 1, which aids in reducing the impact of color channel variations.
    \end{itemize}
 
    \item Distance: this feature is the $L^2$-distance in meters between the positions of the percept and the anchor.
    \item Scale factor: this feature is the scale factor between the sizes of the percept and the anchor.
    \item Time: this feature is the time, in seconds, between the timestamps of the percept and the anchor.
\end{itemize}

 \begin{figure*}[ht!]
\centering
\includegraphics[width=1.0\textwidth]{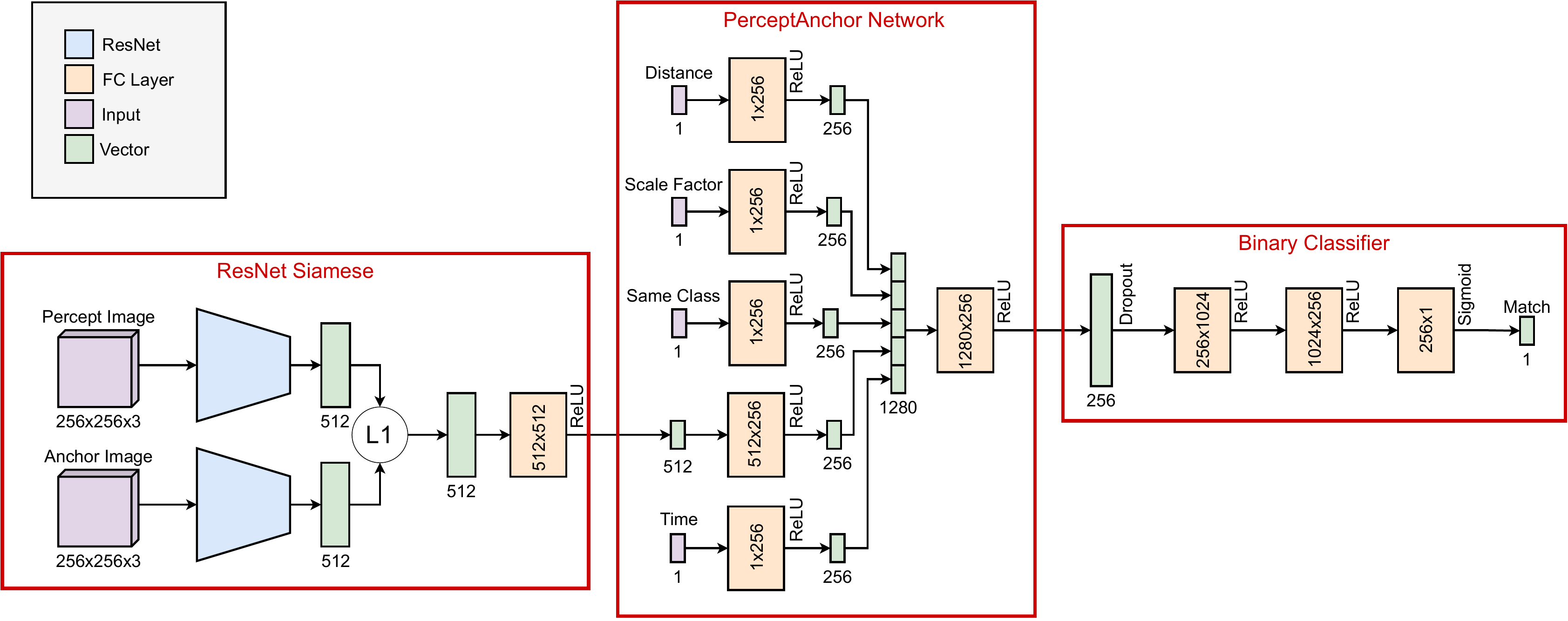}
\caption{\label{fig:sailor_nn} Neural Network used to implement the Matching Function of SAILOR. It is divided into three components: ResNet Siamese, which produces the similarity between two images as a feature vector; PerceptAnchor Network, which encodes each pair of percept-anchor; and Binary Classifier, which classifies each encoded pair as reacquire or acquire.}
\end{figure*}

The neural network used in this work is presented in Figure~\ref{fig:sailor_nn}. This network is composed of three modules:
\begin{itemize}
    \item ResNet Siamese: following the Siamese Convolutional Neural Network used to measure the similarity of two images, we have used two frozen ResNet-18~\cite{targ2016resnet} networks to extract features of the cropped images of each pair of percept-anchor. Then, the $L^1$-distance is applied to the outputs of the ResNet-18. This result is fed into a fully connected layer. As a result, the comparison of the two images from the pair percept-anchor is obtained.
    \item PerceptAnchor Network: this network is in charge of encoding and concatenating the five features that describe each pair percept-anchor. There is a fully connected layer for each input feature (class, ResNet Siamese output, distance, size, timestamps). Then, the outputs are concatenated and fed into another fully connected layer.
    \item Binary Classifier: this network is a Multi-Layer Perceptron (MLP) responsible for classifying the pair percept-anchor as a reacquire. It uses the output of the previous network as its input and returns a value between 0 and 1, thanks to the sigmoid function, that indicates the matching degree between the percept and the anchor. 
\end{itemize}

In the process of training a neural network, choosing the appropriate optimizer and learning rate is crucial to achieving optimal performance. In this study, the training of this neural network was carried out using an Adam optimizer and a learning rate of 0.001. Adam~\cite{kingma2014adam} is a popular optimizer that combines the benefits of two other optimization techniques, namely, Adagrad and RMSprop. This optimizer has been shown to work well in a wide range of deep-learning tasks and is known for its efficient convergence rate. Meanwhile, the learning rate determines the step size taken in each iteration of the optimization process. A low learning rate may cause convergence to be slow, while a high learning rate can lead to overshooting the optimal solution. In this study, a learning rate of 0.001 was chosen based on empirical observations and experimentation.

\begin{figure*}[bp]
\centering
\includegraphics[width=1.0\textwidth, frame]{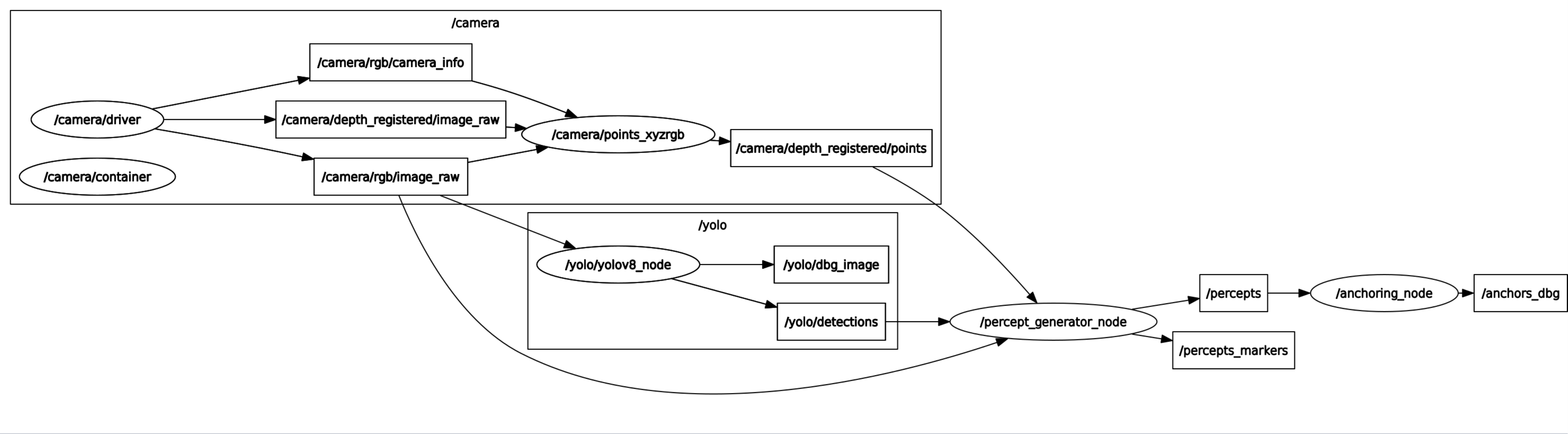}
\caption{\label{fig:rosgraph} Rosgraph of SAILOR, which includes YOLOv8 and camera nodes.}
\end{figure*}

\subsubsection*{Symbolic Layer}
This layer is composed of the knowledge base where the symbolic knowledge is stored. This layer stores the symbolic knowledge as it is proposed in KANT~\cite{kant}. Consequently, the designed knowledge base can be a ROS2 node or a MongoDB. The former means storing the knowledge in memory while the latter provides persistence to the symbolic anchoring system.

The symbolic knowledge employed in KANT is composed of types, objects, predicates and propositions, which is similar to the PDDL elements of the same name. Thus, each real-world object will have a symbolic object and its features will be represented by predicates and propositions.

\section*{Experimental Setup}
\label{sec:experimental_setup}
The experimental setup for this study encompassed four main aspects: SAILOR and ROS 2, Datasets, MERLIN2, and hardware setup. Firstly, the ROS 2 components that compose SAILOR and its communications. Secondly, diverse datasets comprising real-world data were employed to evaluate the performance of the proposed approach. Thirdly, MERLIN2, which is the cognitive architecture where SAILOR is integrated. Lastly, a carefully configured hardware setup was employed.

\subsection*{SAILOR in ROS 2}
SAILOR pipeline has been implemented in ROS 2 for its integration into a real robot called TIAGo. Figure~\ref{fig:rosgraph} shows the rosgraph of SAILOR. It is composed of camera nodes to produce RGB images and point cloud data, the YOLOv8 node, the percept generator node and the anchoring node. The percept generator node is subscribed to RGB images, point cloud data and YOLOv8 detections to produce percepts. Then, the anchoring node is subscribed to these percepts and applies the symbolic anchoring procedures.

\subsection*{Datasets}
The matching function presented in this work is based on a deep learning solution, that is a neural network. As a result, preparing the dataset is a necessary process. In previous machine learning-based works, custom datasets have been created. For instance, in \cite{persson2019semantic} a new dataset is created using a custom labeling tool. This dataset describes each pair of percept-anchor using five similarities (classification, histogram, distance, size, time). However, all data is obtained in scenarios where the robot, a robotic arm, is fixed to a table. Besides, in \cite{gunther2018context} a custom dataset is created using a mobile robot. In this case, pairs of percept-anchor are described using the classification values and distances.

There are several existing datasets that can be used. For instance, KITTI dataset~\cite{Geiger:2013}, is a dataset intended to be used in several tasks such as autonomous driving and object detection. It comprises traffic scenarios recorded with diverse sensor modalities. Another similar dataset is nuScenes~\cite{Caesar:2020} which is a public large dataset for autonomous driving. It contains scenes of images, LIDAR data and ground truth that can be used in several tasks. It includes a Python library to access the data and apply transforms to the positions and sizes of detected objects.

One recent indoor dataset that can be used is MOTFront~\cite{schmauser20223d}. It provides photo-realistic RGB-D images with their corresponding instance segmentation masks, class labels, 3D bounding boxes and 3D poses. The scenes were captured in indoor scenarios with furniture.

\begin{algorithm}[ht!]
\caption{Create dataset algorithm.}\label{alg:create_dataset}
\begin{algorithmic}[1]
\Require $scenes$
\Ensure $pairs$
\State $pairs \gets \emptyset$
\ForEach {$S \in scenes$}
    \State $objects \gets \emptyset$
    \ForEach {$sample \in S$}
        \ForEach {$object \in sample$}
            \ForEach {$o \in objects$}
                \State $pair \gets create\_pair(object, o)$
            \EndFor            
            \State $objects \gets objects \cup object$
        \EndFor
    \EndFor
\EndFor
\Return $pairs$
\end{algorithmic}
\end{algorithm}

With all of this, we have created three new datasets from nuScenes and MOTFront datasets. We have chosen these two datasets to achieve a more diverse solution thanks to the indoor and outdoor data. This would allow us to evaluate the scalability and generality of our learned matching function across different scenarios. 

To create the new datasets, the procedure presented in Algorithm~\ref{alg:create_dataset} is applied to each scene of the datasets. This way, each object of each sample from the scenes is used to create the pairs of percept-anchor. These pairs are created by calculating the five input features of the neural network presented previously.

The previous algorithm is applied to the full MOTFront dataset. In the case of nuScenes, scenes 1, 2, 4, 5, 6, 7, 8, 41, 42  and 43 have been used for training; scenes 3, 12, 13, 14 and 15 for validation; and scenes 1069, 1070, 1071, 1072 and 1073 for testing. Then, both resulting datasets were merged to create the third dataset. The resulting datasets are summed up in Table~\ref{table:datasets}.

\begin{table}[h!]
\centering
\setlength{\tabcolsep}{12pt} 
\renewcommand{\arraystretch}{1.25} 
\setlength{\arrayrulewidth}{1pt} 
\caption{Datasets created using nuScenes and MOTFront data with the number of samples (pairs of percept-anchor) in the training, validation and test splits.}
\vspace{0.25cm}
\label{table:datasets}
\begin{tabular}{|c|c|c|c|} 
 \hline
 \textbf{Dataset} & \textbf{Train} & \textbf{Val} & \textbf{Test} \\[0.5ex]\hline
 \textbf{nuScenes} & 429615 & 33172 & 52806 \\\hline
 \textbf{MOTFront} & 469928 & 104274 & 116655 \\\hline
 \textbf{Mix} & 899543 & 137446 & 169461 \\
 
 \hline
\end{tabular}
\end{table}

\subsection*{MERLIN2}

Cognitive architectures employed in robotics need a way to dynamically obtain knowledge from the environment, which is facing the symbolic anchoring problem. An example of this is presented in \cite{rodriguez2018generating}. It proposes a method for generating symbolic representations of the world from sensory data inside a cognitive architecture. The symbolic representations are stored in the Knowledge Base, nested in a Symbolic Layer, using PDDL~\cite{fox2003pddl2}. Then, that knowledge is used to produce plans that solve the goals of the robot.

In this publication, MERLIN2 architecture~\cite{GONZALEZSANTMARTA2023100477} is used. The architecture is composed of two main systems, Deliberative and Behavioural, that are divided into two layers each. The Deliberative is composed of the Mission Layer and the Planning Layer, where the Knowledge Base can be found. The Behavioural is composed of the Executive Layer and the Reactive Layer, where robot skills can be found.  

\begin{figure}[ht!]
\centering
\includegraphics[width=0.75\textwidth]{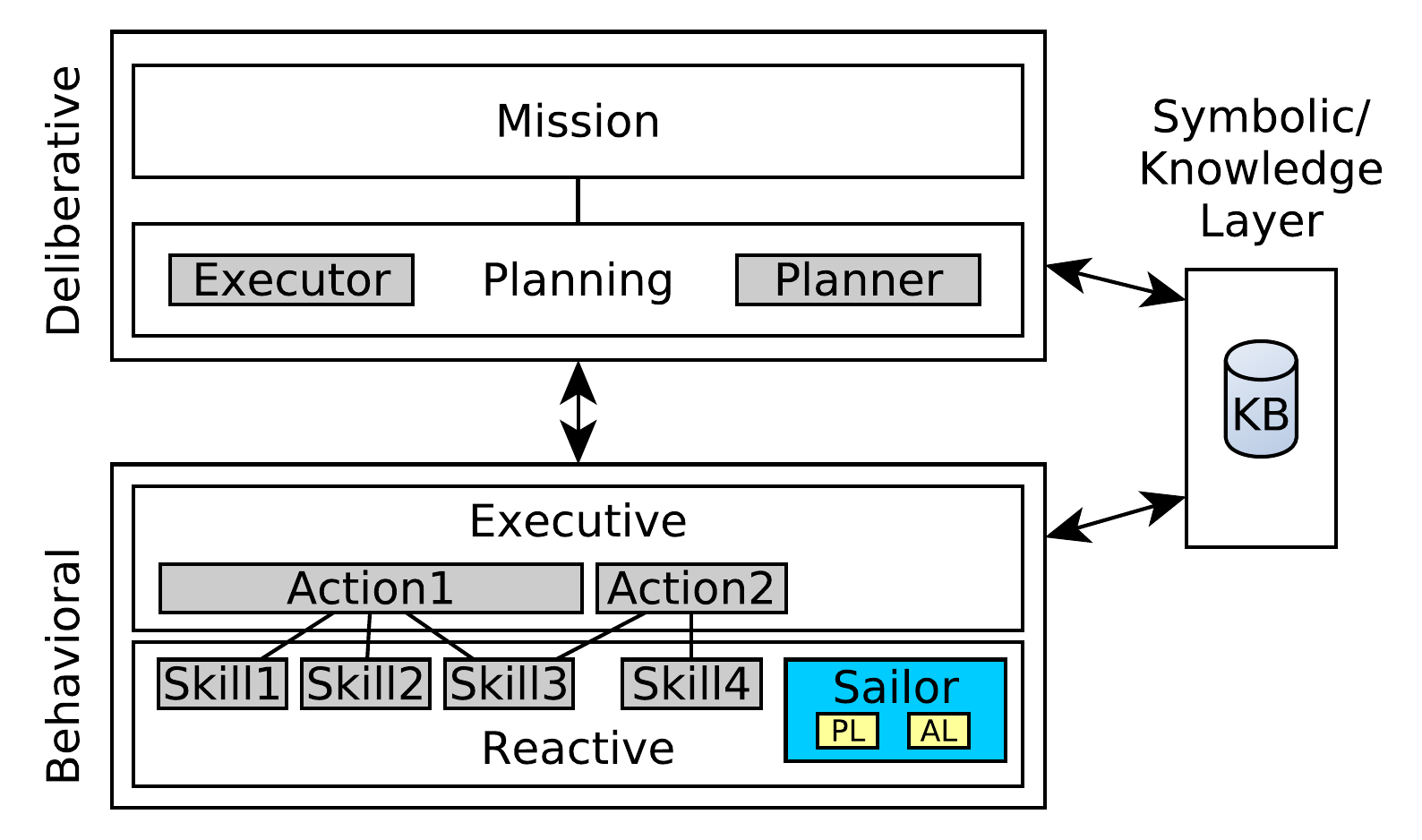}
\caption{\label{fig:MERLIN+SAILOR} MERLIN 2 architecture showing Sailor as a robot skill.}
\end{figure}

Figure~\ref{fig:MERLIN+SAILOR} illustrates how Sailor components are integrated into the MERLIN2 architecture. The sub-symbolic skills, Perceptual Layer (PL) and Anchoring Layer (AL), of SAILOR are integrated into the Reactive Layer from the Behavioral as any other skill available in the robot such as NAV2, Text2Speech, etc. The Symbolic Layer of SAILOR corresponds with the Knowledge Base (KB) of the Planning Layer.

\subsection*{Hardware Setup}

All anchored data were acquired with the use of a TIAGo mobile Robot which is equipped with an Asus Xtion Pro live RGB-D sensor. TIAGo Robot runs ROS Melodic and bridges for interfacing with ROS 2 humble of an external laptop with an Intel(R) i7-8750H CPU, 8 GB RAM and a GTX 1060 Nvidia. Moreover, the training and test were performed using a remote machine with an AMD EPYC 7302P CPU, 256 GB RAM and a Quadro RTX 8000 Nvidia.

\section*{Evaluation}
\label{sec:evaluation}

This evaluation section will discuss the results of the trained models. Specifically, a neural network has been trained on three different datasets to test its ability to anchor symbols to sensory data. The authors examined the performance of the same algorithm, the Matching Function, under the three datasets. The evaluation of the quality of the trained models was carried out by such criteria as a confusion matrix.

\begin{figure*}[ht!]
\centering
\includegraphics[width=1.0\textwidth]{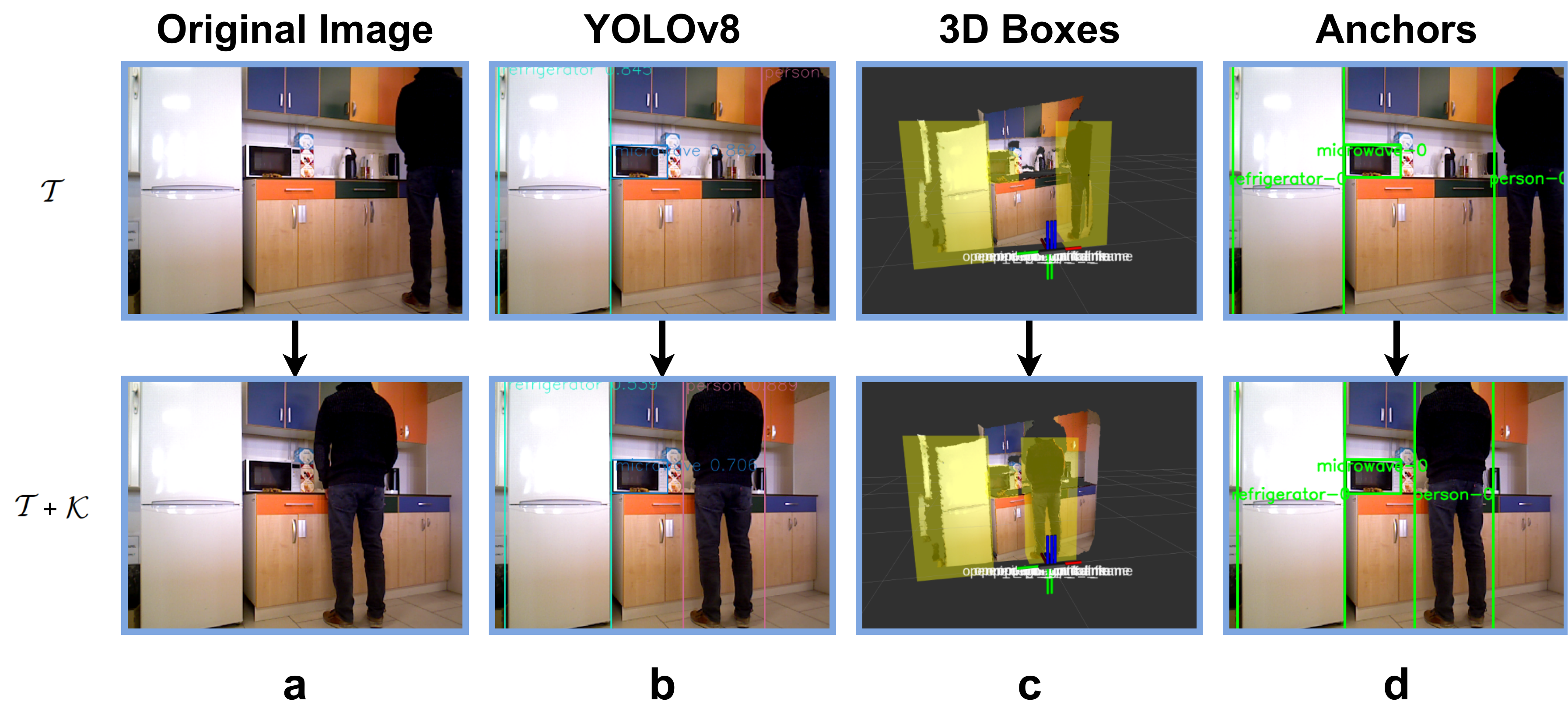}
\caption{\label{fig:sailor_example} Example of using SAILOR twice in a row. From the image captured by the RGB-D camera of the robot (a), YOLOv8 detects certain objects represented by  2D bounding boxes (b). Then, those bounding boxes are converted into 3D Boxes (c) using the point cloud of the camera. The percepts are obtained using the original image and the 3D detected objects. Finally, percepts are acquired or reacquire creating or updating anchors respectively (d).}
\end{figure*}

An example of applying SAILOR is presented in Figure~\ref{fig:sailor_example}. It presents two samples that are taken at different times ($\tau + \kappa$). Figure~\ref{fig:sailor_example} (a) represents the original images gathered by the RGB camera. Then, to the output produced by YOLOv8 (Figure~\ref{fig:sailor_example} (b)), the 3D boxes are calculated (Figure~\ref{fig:sailor_example} (c)). Finally, Figure~\ref{fig:sailor_example} (d) shows the current percepts that have been acquired and reacquired. In this case, there are no new percepts to acquire so all previous percepts are reacquired.

\renewcommand\arraystretch{1.5}
\begin{table}[!ht]
    \centering
    \caption{Generate structure of Confusion Matrix. \label{tab:genericConfusion}}
       
    \begin{tabular}{c >{\bfseries}r @{\hspace{0.7em}}c @{\hspace{0.4em}}c @{\hspace{0.7em}}l}   
        \multirow{10}{*}{\rotatebox{90}{\parbox{1.1cm}{\bfseries\centering actual\\ value}}} & 
          & \multicolumn{2}{c}{\bfseries Prediction outcome} & \\
          & & \bfseries p & \bfseries n & \bfseries total \\
          & p$'$ & \MyBox{True}{Positive} & \MyBox{False}{Negative} & P$'$ \\[2.4em]
          & n$'$ & \MyBox{False}{Positive} & \MyBox{True}{Negative} & N$'$ \\
          & total & P & N &
        \end{tabular}
\end{table}

Table~\ref{tab:genericConfusion} illustrates the template of the confusion matrix. Confusion matrix~\cite{categorical1998glossary} compares the predicted labels of a model with the true labels of the data it was trained on. The matrix contains four different values: true positives (TP), the number of sensory data  that our model correctly anchors and are positive; false positives (FP), 
the number of sensory data incorrectly anchored and are positive; true negatives (TN), the number of sensory data that our model does not anchor and does not need to be anchored; and false negatives (FN), the sensory data that our model does not anchor and it does need to be anchored.
Afterward, these values are then used to calculate different evaluation metrics, such as accuracy, precision, recall and F1-score. Accuracy measures the proportion of correct predictions, while precision measures the proportion of true positives among all positive predictions. Recall measures the proportion of true positives among all actual positives, while the F1-score is the harmonic mean of precision and recall.

Tables~\ref{fig:NUSCENEconfusion}, \ref{fig:MOTFRONTconfusion}, \ref{tab:mixedconfusion} and \ref{tab:MIXEDconfusion} give the confusion matrix of symbolic anchoring accuracy in times and percentages for the testing data from the three datasets: nuScenes, MOTFront and Mix.

Table~\ref{fig:NUSCENEconfusion} shows the confusion matrices associated with our matching function trained with nuScenes and tested with MOTFront and nuScenes groups. The findings, in this case, present a non-common behavior. nuScenes behavior is better with MOTFront test data than with nuScenes itself in True Positive values. It achieves 97.94\% against 89.77\%. True Negative values present a rate of 96.18\% in MOTFront and a rate of 99.86\% in nuScenes.

\renewcommand\arraystretch{1.5}
\begin{table}[!ht]
    \caption{Confusion matrix for the model trained on the nuScenes dataset.\label{fig:NUSCENEconfusion}}
\begin{minipage}{0.45\textwidth}
\centering
    \begin{tabular}{c >{\bfseries}r @{\hspace{0.7em}}c @{\hspace{0.4em}}c @{\hspace{0.7em}}l}   
    \multirow{10}{*}{\rotatebox{90}{\parbox{1.1cm}{\bfseries\centering actual\\ value}}} &        
            & \multicolumn{2}{c}{\bfseries Prediction outcome} & \\
          & & \bfseries p & \bfseries n & \bfseries total \\
         & p$'$ & \MyBox{33397}{97.94\%} & \MyBox{3152}{3.82\%} & 36549 \\[2.4em]
          & n$'$ & \MyBox{702}{2.06\%} & \MyBox{79404}{96.18\%} & 80106 \\
          & total & 34099 & 82556 &\\ 
             & \multicolumn{4}{c}{\bfseries Model tested on the MOTFront dataset} 
    \end{tabular}
\end{minipage}
\begin{minipage}{0.45\textwidth}
\centering
\begin{tabular}{c >{\bfseries}r @{\hspace{0.7em}}c @{\hspace{0.4em}}c @{\hspace{0.7em}}l}   
           \multirow{10}{*}{\rotatebox{90}{\parbox{1.1cm}{\bfseries\centering ~\\ 
          ~}}} &
            & \multicolumn{2}{c}{\bfseries Prediction outcome} & \\
          & & \bfseries p & \bfseries n & \bfseries total \\
          & p$'$ & \MyBox{1685}{89.77\%} & \MyBox{104}{0.20\%} & 1789 \\[2.4em]
          & n$'$ & \MyBox{192}{10.23\%} & \MyBox{50825}{99.80\%} & 51017 \\
          & total & 1877 & 50929 & \\
            & \multicolumn{4}{c}{\bfseries Model tested on the nuScenes dataset} 
        \end{tabular}
\end{minipage}

\end{table}

Table \ref{fig:MOTFRONTconfusion} shows the confusion matrices associated with our matching function trained with MOTFront and tested with MOTFront and nuScenes groups. The results present several similar results in True Positive and MOTFront achieving 99.78\%, this is 0.92 \% better than using nuScenes. True Negative values present a rate of 99.78\% in MOTFront and achieve a rate of 98.86\% in nuScenes. 

\renewcommand\arraystretch{1.5}
\begin{table}[!ht]
    \caption{Confusion matrix for the model trained on the MOTFront dataset.\label{fig:MOTFRONTconfusion}}
\begin{minipage}{0.45\textwidth}
    \begin{tabular}{c >{\bfseries}r @{\hspace{0.7em}}c @{\hspace{0.4em}}c @{\hspace{0.7em}}l}   
    \multirow{10}{*}{\rotatebox{90}{\parbox{1.1cm}{\bfseries\centering actual\\ value}}} & 
            & \multicolumn{2}{c}{\bfseries Prediction outcome} & \\
          & & \bfseries p & \bfseries n & \bfseries total \\
           & p$'$ & \MyBox{36370}{99.81\%} & \MyBox{179}{0.22\%} & 36549 \\[2.4em]
          & n$'$ & \MyBox{69}{0.19\%} & \MyBox{80037}{99.78\%} & 80106 \\
          & total & 36439 & 80216 &\\
             & \multicolumn{4}{c}{\bfseries Model tested on the MOTFront dataset} 
    \end{tabular}
\end{minipage}
\begin{minipage}{0.45\textwidth}
    \begin{tabular}{c >{\bfseries}r @{\hspace{0.7em}}c @{\hspace{0.4em}}c @{\hspace{0.7em}}l}   
        \multirow{10}{*}{\rotatebox{90}{\parbox{1.1cm}{\bfseries\centering ~\\ 
          ~}}} &
            & \multicolumn{2}{c}{\bfseries Prediction outcome} & \\
          & & \bfseries p & \bfseries n & \bfseries total \\
           & p$'$ & \MyBox{1201}{100.00\%} & \MyBox{588}{1.14\%} & 1789 \\[2.4em]
          & n$'$ & \MyBox{0}{0.00\%} & \MyBox{51017}{98.86\%} & 51017 \\
          & total & 1201 & 51605 &\\
            & \multicolumn{4}{c}{\bfseries Model tested on the nuScenes dataset} 
    \end{tabular}   
\end{minipage}
\end{table}

Table \ref{tab:mixedconfusion} shows the confusion matrices associated with our matching function trained with the Mix dataset and tested with MOTFront and nuScenes groups. The trained model presents better True Positive values in the MOTFront test data than with nuScenes. It achieves 99.76\% against 95.00\%. True Negative values present a rate of 99.03\% in MOTFront and 99.83\% in nuScenes. 
 
\renewcommand\arraystretch{1.5}
\begin{table}[!ht]
    \caption{Confusion matrix for the model trained on the Mix dataset.\label{tab:mixedconfusion}}     
\begin{minipage}{0.45\textwidth}
    \begin{tabular}{c >{\bfseries}r @{\hspace{0.7em}}c @{\hspace{0.4em}}c @{\hspace{0.7em}}l}   
        \multirow{10}{*}{\rotatebox{90}{\parbox{1.1cm}{\bfseries\centering actual\\ value}}} & 
          
            & \multicolumn{2}{c}{\bfseries Prediction outcome} & \\
          & & \bfseries p & \bfseries n & \bfseries total \\
          & p$'$ & \MyBox{1292}{95.00\%} & \MyBox{497}{0.97\%} & 1789 \\[2.4em]
          & n$'$ & \MyBox{68}{5.00\%} & \MyBox{50949}{99.03\%} & 51017 \\
          & total & 1360 & 51446 &\\ 
             & \multicolumn{4}{c}{\bfseries Model tested on the MOTFront dataset} 
        \end{tabular}
\end{minipage}
\begin{minipage}{0.45\textwidth}
       \begin{tabular}{c >{\bfseries}r @{\hspace{0.7em}}c @{\hspace{0.4em}}c @{\hspace{0.7em}}l}   
           \multirow{10}{*}{\rotatebox{90}{\parbox{1.1cm}{\bfseries\centering ~\\ 
          ~}}} &
            & \multicolumn{2}{c}{\bfseries Prediction outcome} & \\
          & & \bfseries p & \bfseries n & \bfseries total \\
           & p$'$ & \MyBox{36411}{99.76\%} & \MyBox{138}{0.17\%} & 36549 \\[2.4em]
          & n$'$ & \MyBox{88}{0.24\%} & \MyBox{80018}{99.83\%} & 80106 \\
          & total & 36499 & N 80156 \\
            & \multicolumn{4}{c}{\bfseries Model tested on the nuScenes dataset} 
        \end{tabular}
\end{minipage}
\end{table}

Finally, table \ref{tab:MIXEDconfusion} shows the confusion matrix associated with our dataset, with data from both datasets (nuScenes and MotFront). True Positive values present a rate of 99.59\% and True Negative values present a rate of 99.52\%. 

\renewcommand\arraystretch{1.5}
    \begin{table}[ht]
    \centering
    \caption{Confusion Matrix for the model trained and tested on the Mix dataset.\label{tab:MIXEDconfusion}}
        
       \begin{tabular}{c >{\bfseries}r @{\hspace{0.7em}}c @{\hspace{0.4em}}c @{\hspace{0.7em}}l}   
          \multirow{10}{*}{\rotatebox{90}{\parbox{1.1cm}{\bfseries\centering actual\\ value}}} & 
            & \multicolumn{2}{c}{\bfseries Prediction outcome} & \\
          & & \bfseries p & \bfseries n & \bfseries total \\
           & p$'$ & \MyBox{37703}{99.59\%} & \MyBox{635}{0.48\%} & 38338 \\[2.4em]
          & n$'$ & \MyBox{156}{0.41\%} & \MyBox{130967}{99.52\%} & 131123 \\
          & total & 37859 & 131602 &
        \end{tabular}
    \end{table}

As a result of previous findings, Table \ref{tab:Accuracyetal} presents average symbolic anchoring classification accuracy, precision, recall and F1-score. The accuracy metric defines the total number of correctly classified data over the total number of data. However, this metric is not enough for non-balanced datasets. Precision defines how accurate is the model out of those predicted positives and measures how many of them are actually positives. This metric is a good measure to determine when the cost of a False Positive is high, this means measuring how many elements are we detecting wrongly. Recall defines how many of the Actual Positives our model captures by labeling it as Positive (True Positive). This metric helps us to select the best model when there is a high cost associated with False Negative. In this case, when we do not acquire an anchor that is present in the scene. F1-Score is a metric to be applied if it is needed to seek a balance between Precision and Recall. In these large datasets would be an uneven class distribution (a considerable number of certain Negatives).

Results present high accuracy rates and the enhancements are close to insignificant to previous results.  Taking a look at recall we can note differences in nuScenes behavior, increasing MOTFront performance recognizing nuScenes elements and this improvement is also presented in the F1-score. 

Comparing these findings it seems that MotFront has a high influence in the Mix Dataset and the classification of our matching function is highly influenced to better classify the candidate anchors from this dataset.

\begin{table}[ht]
 \centering

    \caption{Resulting average classification accuracy together with precision, Recall and F1-score for each model tested in our approach to performing the symbolic anchoring functionalities.\label{tab:Accuracyetal}}
    \vspace{0.25cm}

    \begin{tabular}{|P{1.5cm}|P{1.5cm}|P{1.5cm}|P{1.5cm}|P{1.5cm}|}
        \hline
         &Accuracy & Precision  & Recall  & F1-score  \\ \hline

          \multicolumn{5}{|c|}{\bfseries nuScenes Dataset}  \\ \hline
        MOTFront & 0.967 & 0.979 & 0.914 & 0.945 \\ \hline
        nuScenes & 0.994 & 0.898 & 0.942 & 0.919 \\ \hline \hline
          
          \multicolumn{5}{|c|}{\bfseries MOTFront Dataset} \\ \hline
        MOTFront & 0.998 & 0.998 & 0.995 & 0.997 \\ \hline
        nuScenes & 0.989  & 1.000 & 0.671 & 0.803 \\ \hline \hline

          \multicolumn{5}{|c|}{\bfseries Mix Dataset}  \\ \hline       
        MOTFront & 0.998 & 0.998 & 0.996 & 0.997 \\ \hline
        nuScenes & 0.989 & 0.950 & 0.722 & 0.821 \\ \hline
        Mix Test & 0.995 & 0.996 & 0.983 & 0.990 \\ \hline
    \end{tabular}
\end{table}

\subsection*{Contribution evaluation}

This research proposed three main contributions that have been validated: 
\begin{enumerate}
    \item It presented an updated symbolic anchoring pipeline based on state-of-the-art works. The presented pipeline is based on object detection followed by percept generation, which implies using the bounding boxes to get physical features from the point cloud of the camera.
    \item The matching function based on deep learning for an anchoring system achieves more than 96\% accuracy in all cases tested.
    \item The evaluation of datasets and models for symbolic anchoring is being validated by mixing nuScenes (outdoors) and MOTFront (indoors) datasets. The resulting dataset is available at Hugging Face\footnote{\url{https://huggingface.co/datasets/unileon-robotics/sailor}}.
    \item A set of ROS 2 components have been validated and tested in a real robotic Platform TIAGo. These components are publicly available in a GitHub Repository\footnote{\url{https://github.com/MERLIN2-ARCH/sailor}}.
\end{enumerate}

\section*{Conclusions}
\label{sec:conclusions}

In this work, we have developed SAILOR, a set of software components for ROS 2 to provide a symbolic anchoring process in a cognitive architecture of a robotic system. We have further introduced a symbolic anchoring pipeline that performs an object detection process and afterward performs a point cloud-based physical feature extraction process, rather than the reviewed state-of-the-art works that perform the process all the way around. 

The developed matching function is based on a neural network. It defines a ResNet Siamese followed by a PerceptAnchor Network and finishing in a Binary Classifier. The use of state-of-the-art datasets allowed us to measure its performance. The classifier performance for the nuScenes dataset (outdoor environment) shows that the nuScenes test had better performance in accuracy and recall than the MOTFront test appears to perform better overall in the outdoor environment and the MOTFront test has higher precision and F1-score than the nuScenes test model.
For the MOTFront dataset (indoor environment), the MOTFront test appears to have a better precision score, as expected, while the Mix dataset has a better recall.
Overall, it's difficult to determine which model is definitively better for either indoor or outdoor environments, as the performance can depend on many factors such as the specific task, dataset and evaluation metrics. However, based on the provided table, we can say that the Mix dataset appears to perform well in both indoor and outdoor environments, while the MOTFront dataset has a high precision score for indoor environments.

Finally, it was initially tested and deployed in the TIAGo robot. However, further, development is needed. The reason is that the robot works with a noisy sensor and the SAILOR performance is directly linked to the slower component in the pipeline. Initial tests with noisy network connections reduce the process performance.

\bibliography{main}



\section*{Acknowledgments}
This work has been partially funded by an FPU fellowship provided by the Spanish Ministry of Universities (FPU21/01438) and the Grant PID2021-126592OB-C21 funded by MCIN/AEI/10.13039/501100011033 and by ERDF A way of making Europe.

\section*{Declarations}

\noindent
\textbf{Conflict of interest} The authors assert no conflicts of interest in this work.\\

\noindent
\textbf{Ethics approval} The authors confirm that they have complied with the publication ethics and state that this work is original and has not been used for publication anywhere before.\\



\noindent
\textbf{Availability of data and materials} The data is available on Hugging Face (\url{https://huggingface.co/datasets/unileon-robotics/sailor}).\\

\noindent
\textbf{Code availability}  The data is available on GitHub (\url{https://github.com/MERLIN2-ARCH/sailor}).\\

\noindent
\textbf{Authors' contributions} All authors contributed to the study conception and design. Data and code preparation were performed by Miguel Á. González-Santamarta. The first draft of the manuscript was written by Miguel Á. González-Santamarta and all authors commented on previous versions of the manuscript. All authors read and approved the final manuscript.

\end{document}